\documentclass[letterpaper]{article} 
\usepackage{aaai24}  
\usepackage{times}  
\usepackage{helvet}  
\usepackage{courier}  
\usepackage[hyphens]{url}  
\usepackage{graphicx} 
\usepackage{multirow}
\usepackage{xcolor}
\usepackage{amsmath,amsfonts}
\usepackage{subcaption}
\usepackage{natbib}  
\usepackage{caption} 
\usepackage{algorithm}
\usepackage{algorithmic}

\usepackage{newfloat}
\usepackage{listings}
\DeclareCaptionStyle{ruled}{labelfont=normalfont,labelsep=colon,strut=off} 
\floatstyle{ruled}
\newfloat{listing}{tb}{lst}{}
\floatname{listing}{Listing}
\title{Tuning Large language model for End-to-end Speech Translation}
\author {
    Hao Zhang\textsuperscript{\rm 1},
    Nianwen Si\textsuperscript{\rm 1,2},
    Yaqi Chen\textsuperscript{\rm 1},
    Wenlin Zhang\textsuperscript{\rm 1}, \\
    Xukui Yang\textsuperscript{\rm 1},
    Dan Qu\textsuperscript{\rm 1},
    Xiaolin Jiao\textsuperscript{\rm 3},
}
\affiliations {
    \textsuperscript{\rm 1}University of Information Engineering, ZhengZhou 450000, China\\
    \textsuperscript{\rm 2}Department of Electronic Engineering, Tsinghua University, Beijing 100084, China\\
    \textsuperscript{\rm 3}Zhengzhou University, ZhengZhou 450000, China\\
    \{haozhang0126, snw1608, chyaqi163, wenlinzzz, gzyangxk, qudan\_xd\}@163.com \\
    jxl5056@sina.com
}

\usepackage{bibentry}

\begin{document}

\maketitle

\begin{abstract}
With the emergence of large language models (LLMs), multimodal models based on LLMs have demonstrated significant potential. Models such as LLaSM, X-LLM, and SpeechGPT exhibit an impressive ability to comprehend and generate human instructions. However, their performance often falters when faced with complex tasks like end-to-end speech translation (E2E-ST), a cross-language and cross-modal translation task. In comparison to single-modal models, multimodal models lag behind in these scenarios. This paper introduces LST, a \textbf{L}arge multimodal model designed to excel at the E2E-\textbf{ST} task. LST consists of a speech frontend, an adapter, and a LLM backend. The training of LST consists of two stages: (1) Modality adjustment, where the adapter is tuned to align speech representation with text embedding space, and (2) Downstream task fine-tuning, where both the adapter and LLM model are trained to optimize performance on the E2E-ST task. Experimental results on the MuST-C speech translation benchmark demonstrate that LST-13B achieves BLEU scores of 30.39/41.55/35.33 on En-De/En-Fr/En-Es language pairs, surpassing previous models and establishing a new state-of-the-art. Additionally, we conduct an in-depth analysis of single-modal model selection and the impact of training strategies, which lays the foundation for future research. We will open up our code and models after review.
\end{abstract}

\section{Introduction}
End-to-end speech translation (E2E-ST) \cite{weiss2017sequence,berard2018end,sperber2019attention} refers to translate the speech in source language into text in target language without generating the intermediate transcription. It has emerged as a prominent paradigm due to its advantages over the cascade model, such as lower latency and less error progradation \cite{liu2020bridging,dong2021listen}. However, effectively training such a model remains a challenge due to the task complexity and limited data availability. To attain optimal performance, previous studies have introduced sophisticated techniques, such as knowledge distillation \cite{liu2019end,gaido2020end,gaido2020knowledge,inaguma-etal-2021-source,zhao-etal-2021-mutual}, contrastive learning \cite{ye2022cross,ouyang2022waco,Zhang2023ImprovingST}, data augmentation \cite{bahar-etal-2019-using,lam-etal-2022-sample,zhao-etal-2023-generating,fang2023back,zhang2023dub}, mixup \cite{fang-etal-2022-stemm,cheng2023m}, to assist training. As a result, E2E-ST models have achieved almost comparable or even superior performance than the conventional cascade ST models \cite{ye2022cross}. Nevertheless, it appears that the performance of the current E2E-ST model has encountered certain limitations. This is attributed to the fact that most of these methods require training from scratch \cite{xu-etal-2021-stacked} or start from a weak initialization \cite{6acdf73253e64b7eb908c3a90c13e2d9,tang-etal-2021-improving,fang-etal-2022-stemm}.

Recently, large language models (LLMs) have experienced rapid advancements \cite{thoppilan2022lamda,touvron2023llama}. With the increase of data and parameter scale, LLMs have demonstrated many impressive capabilities, including in-context learning, instruction following and Chain-of-Thought \cite{chowdhery2022palm}. Simple tweaks on downstream tasks can even approach or surpass a well-designed small models. A fundamental assumption in deep learning is that different modalities share the same semantic space \cite{merullo2022linearly}. Building upon this assumption, the next natural step is to develop the cascaded multimodal models, which aims to transfer the ability of LLM to other modalities. In the field of computer vision, there has been a lot of relevant researches, such as MiniGPT-4 \cite{zhu2023minigpt}, LLaVA \cite{liu2023visual} and LLaMA-adapter \cite{zhang2023llama,gao2023llama}, which have shown extremely superior performance on task like Science QA. However, in the field of speech, although models X-LLM \cite{chen2023x} and SpeechGPT \cite{zhang2023speechgpt} are capable of ‘listening’ and ‘speaking’, there still exists a significant performance gap compared to the best single modality model in complex downstream task \cite{chen2023x}.

\begin{figure*}[ht]
\centering
\includegraphics[width=0.6\textwidth]{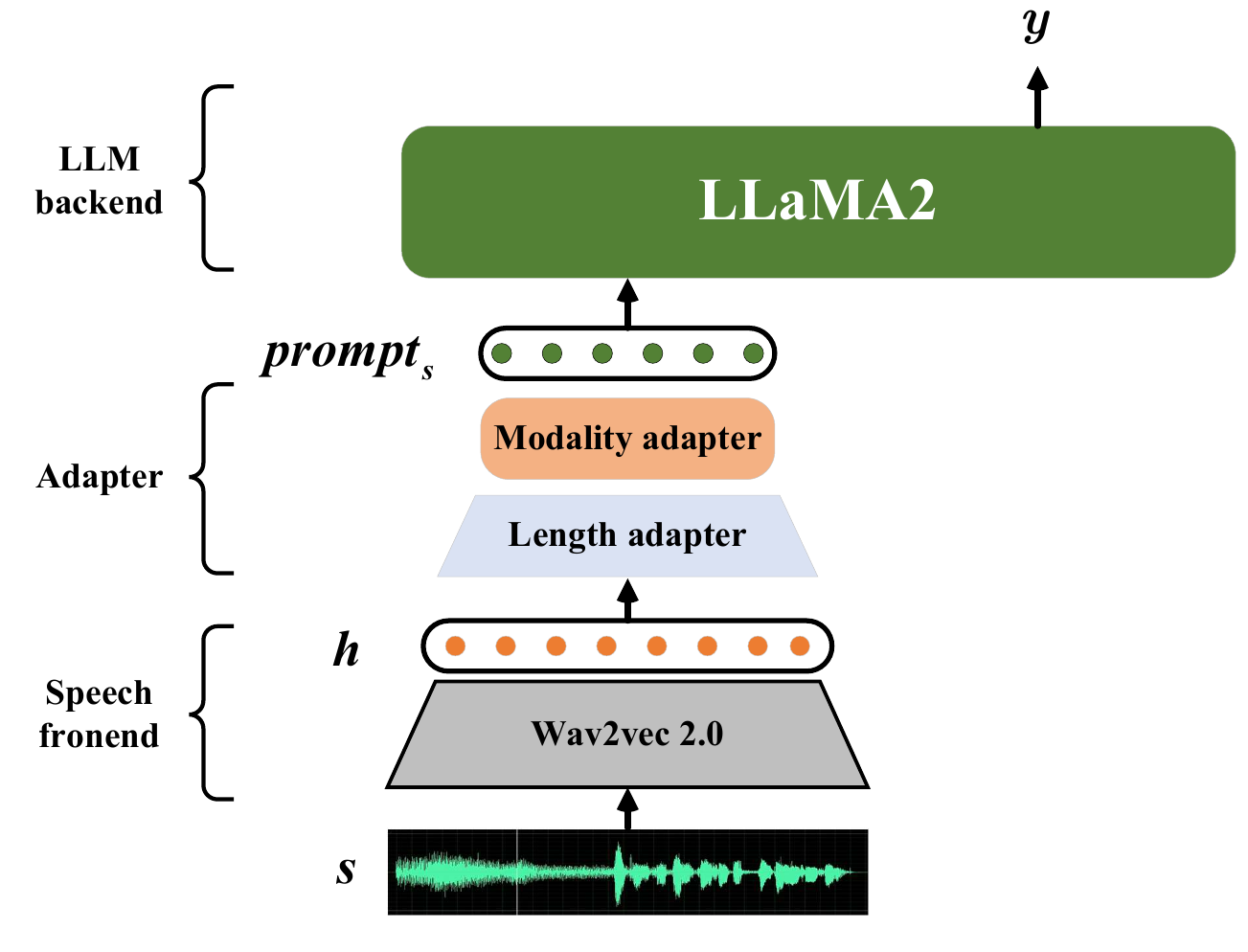} 
\caption{The overall architecture of LST. It contains three modules, a speech frontend, an adapter, and a LLM backend. Given the source speech $s$, we utilize the speech frontend to extract representation $h = frontend(s)$. And then we feed it into the adapter to get the soft prompt  $promp{t_s} = adapter({h_s})$, which is further used to prompt the LLM to generate the text translation $y = LLM(promp{t_s})$  in the target language.}
\label{main_figure}
\end{figure*}

To this end, we propose LST, a \textbf{L}arge multimodal model that excels at E2E-\textbf{ST} task. Following the previous works \cite{zhu2023minigpt,chen2023x,liu2023visual}, LST adopts the cascade strategy to improve scalability and facilitate the utilization of advanced research results in single modality. It comprises a speech frontend, an adapter, and a LLM backend. We first extract the speech representations through the speech frontend, and then transform them into the text embedding space using the lightweight adapter. The transformed representation serves as the soft prompt for the LLM backend, prompting it to generate the corresponding ground-truth text translation in the target language. The training of LST involves two stages. (1) Modality adjustment. In this stage, we solely tune the adapter to align the speech representation with the text embedding space. (2) Downstream task fine-tuning. We freeze the parameter of the speech frontend, and train both the adapter and LLM model to optimize the performance on the E2E-ST task.

Experiments on the MuST-C \cite{di2019must} datasets show that LST-13B achieves 30.39/41.55/33.73 BLEU scores on En-De/En-Fr/En-Es language pairs, significantly surpassing the previous models, achieving state-of-the-art performance. Moreover, we undertake a systematic analysis and experimentation, revealing the following key findings: (1) The necessity of two-stage training: Directly attempting to finetune the entire model in an end-to-end manner, without the initial stage, yields notably poor results. (2) Enhanced efficacy of ST datasets over automatic speech recognition (ASR) datasets in the first stage: The utilization of ST datasets produces marginally superior outcomes. (3) Significance of linguistic supervision within the speech frontend: The presence of linguistic supervision significantly influences final performance. Particularly, CTC-finetuned Wav2vec 2.0 \cite{baevski2020wav2vec} yields substantial improvements over the original Wav2vec 2.0 model. This enhancement can be attributed to the constrained modality adjustment capability of the lightweight adapter in the first stage. (4) Backend selection: Opting for LLM after supervised fine-tuning (SFT) is not the most optimal choice. Interestingly, when only the first training stage is executed, during which the LLM parameters remain fixed, LLM after SFT exhibits improved results. However, for the comprehensive two-stage training, the original foundational LLM performs better.  

Our contributions are summarized as follows:
\begin{itemize}
\item We propose LST, a large multimodal model constructed from cascaded single-modality models that excels in E2E-ST task.
\item We systematically studied the impacts of training strategies and sub-module selections within LST, encompassing both the speech frontend and the LLM backend. This comprehensive analysis lays the foundation for future research endeavors.
\item We conduct experiments on the MuST-C dataset and achieve new state-of-the-art performance. At the same time, because of its simple structure and training method, the proposed approach can serve as a novel baseline for subsequent researches. We have made the complete source code and model checkpoint for this study publicly available.
\end{itemize}

\section{Methods}

We begin with the fundamental problem formulation of E2E-ST. Then, the structure and the training strategy of LST is introduced in detail. 

\subsection{Problem Formulation of E2E-ST}
\label{table1}
The speech translation corpus usually contains speech-transcription-translation triples, denoted as ${D_{ST}}=\{ s,x,y\} $, where $s$,$x$,$y$ are the audio, the corresponding transcription in the source language, and the translation in the target language, espectively. E2E-ST aims to generate the translation $y$ directly without generating the intermediate transcription. 

\subsection{Model Architecture}
The objective of LST is to leverage the powerful understanding capabilities of LLM, in conjunction with existing single-modality research achievements, to translate speech into text. The overall architecture is illustrated in Figure \ref{main_figure}. LST comprises three modules: (a) speech frontend, responsible for extracting semantic representation from the input speech signal; (b) adapter, which encompasses both length and modality adapters. The extracted representation typically exceeds a length of 500 \cite{dong2021listen}. We use the length adapter, a 1-dimensional convolution, to reduce the length. The modality adapter is a simple linear layer that transforms speech representation into text embedding space. We leave the exploration of potentially more effective and sophisticated architectural designs for future work; (c) LLM backend, where the transformed representation serves as a soft prompt for the LLM \cite{merullo2022linearly}, prompting it to generate the corresponding translation in the target language.

\begin{table}[]
\centering
\begin{tabular}{c|c|cc}
\hline
                            & language & hours & \#sents \\ \hline
\multirow{3}{*}{ST(MuST-C)} & En-De    & 408   & 234K   \\
                            & En-Fr    & 492   & 280K   \\
                            & En-Es    & 504   & 270K   \\ \hline
ASR(Librispeech)            & En       & 960   & 280K   \\ \hline
\end{tabular}
\caption{Statistics of all datasets. The duration of each sample in Librispeech is about 10s, while the duration of samples in MuST-C varies greatly. Therefore, although the total durations of Librispeech and MuST-C is different, the sample numbers are relatively close.}
\label{table_data}
\end{table}

\subsection{Training}

The entire training is divided into two stages, namely modality adjustment and downstream task fine-tuning.

\noindent\textbf{First Stage Training.} At this stage, we keep the speech frontend and the LLM backend frozen, and only the adapter parameters are trainable. The main objectives of this stage are as follows: (1) speech length reduction. (2) modality transformation of speech representation. For this stage, the choice of training data is flexible and can include either ASR or ST datasets. The ASR dataset is denoted as ${D_{ASR}} = \{ s,x\} $. When using ASR dataset, the loss is defined as: 

\begin{equation}
{\cal L}(\theta ) =  -\sum\limits_{(s,x) \in {D_{ASR}}} {\log {p_\theta }(x|s)}
\label{equation:1}
\end{equation}

\noindent{When} using ST dataset, the loss is defined as:

\begin{equation}
{\cal L}(\theta ) =  -\sum\limits_{(s,y) \in {D_{ASR}}} {\log {p_\theta }(y|s)}
\label{equation:1}
\end{equation}

\noindent{where} $\theta {\rm{ = \{ adapter\} }}$ is the trainable parameters.

\noindent\textbf{Second Stage Training.} Previous studies have shown that making the parameters of the single-modality frontend trainable has minimal impact \cite{merullo2022linearly}. Therefore, in this stage, in this stage, we freeze the speech frontend and focus on training both the adapter and the LLM backend using the foundations established in the first stage. The objective of this stage is to enhance the model's capabilities for specific downstream tasks. The loss calculation remains the same as in the first stage, with the only difference being that the trainable parameter becomes $\theta {\rm{ = \{ adapter,LLM\} }}$.

\section{Experiments}

\subsection{Dataset and Processing}
MuST-C\footnote{https://mt.fbk.eu/must-c-releases/} \cite{di2019must} a multilingual dataset extracted from TED talks, including source speech, transcriptions, and text translations. Its source language is English, and the target language cover eight language direction. It is one of the most extensive training data for speech translation. We conduct experiments on the English-German, English-French and English-Spanish language pairs. We use dev set for validation and tst-COMMON set for test. When ASR dataset is used in the first stage, we use 960 hours of data from Librispeech \cite{panayotov2015librispeech}. The detailed statistics
of all datasets included are shown in Table \ref{table_data}.

We use the original 16-bit 16 kHz mono-channel audio waveform as speech input \cite{fang-etal-2022-stemm}. We true case all translation texts.

\begin{table*}[ht]
\centering
\begin{tabular}{l|cc|cccc}
\hline
\multirow{2}{*}{Models} & \multicolumn{2}{c|}{Self-supervised model} & \multicolumn{4}{c}{BLEU}                         \\ \cline{4-7} 
                        & Speech             & Text             & En-De & En-Fr & \multicolumn{1}{l|}{En-Es} & Avg \\ \hline
Fairseq-ST \cite{wang2020fairseq}               & $\times$           & $\times$         & 22.7  & 32.9  & \multicolumn{1}{l|}{27.2}  & 27.6 \\
STAST \cite{liu2020bridging}                  & $\times$           & $\times$         & 23.06 & $-$   & \multicolumn{1}{l|}{$-$}   & $-$    \\
SATE \cite{xu-etal-2021-stacked}                   & $\times$           & $\times$         & 28.1  & $-$   & \multicolumn{1}{l|}{$-$}   & $-$    \\
TDA \cite{du2022regularizing}                    & $\times$           & $\times$         & 27.1  & 37.4  & \multicolumn{1}{l|}{$-$}   & $-$   \\
Chimera \cite{6acdf73253e64b7eb908c3a90c13e2d9}                & $\checkmark$       & $\times$         & 26.3  & 35.6  & \multicolumn{1}{l|}{30.6}  & 30.8   \\
ConST  \cite{ye2022cross}                 & $\checkmark$       & $\times$         & 28.3  & 38.3  & \multicolumn{1}{l|}{31.0}  & 32.5  \\
FCCL  \cite{Zhang2023ImprovingST}                 & $\checkmark$       & $\times$         & 29.0  & 38.3  & \multicolumn{1}{l|}{31.9}  & 33.1  \\
STEMM \cite{fang-etal-2022-stemm}                  & $\checkmark$       & $\times$         & 28.7  & 37.4  & \multicolumn{1}{l|}{31.0}  & 32.4 \\
M$^3$ST \cite{cheng2023m}                 & $\checkmark$       & $\times$         & 29.3  & 38.5  & \multicolumn{1}{l|}{32.4}  & 33.4 \\
CRESS \cite{fang2023understanding}                  & $\checkmark$       & $\times$         & 29.4  & 40.1  & \multicolumn{1}{l|}{33.2}  & 34.2 \\
MSP \cite{zhang2023improving}                     & $\checkmark$       & $\checkmark$     & 30.2  & $-$   & \multicolumn{1}{l|}{$-$}   & $-$ \\
STPT$^\dag$  \cite{tang2022unified}            & $\checkmark$       & $\checkmark$     & $-$   & 39.7  & \multicolumn{1}{l|}{33.1}  & $-$   \\
SpeechUT$^\dag$ \cite{zhang2022speechut}        & $\checkmark$       & $\checkmark$     & 30.1  & 41.4  & \multicolumn{1}{l|}{33.6}  & 35.0 \\ \hline
LST-7B                  & $\checkmark$       & $\checkmark$     & 29.15 &  40.77     & \multicolumn{1}{l|}{33.07}      &   34.33  \\  
LST-13B                 & $\checkmark$       & $\checkmark$     & \textbf{30.39}      &   \textbf{41.55}    & \multicolumn{1}{l|}{\textbf{35.33}}      &  \textbf{35.75}   \\ \hline
\end{tabular}
\caption{BLEU scores on MuST-C tst-COMMON set. "Speech" denotes using the speech SSL model, such as Wav2vec 2.0 \cite{baevski2020wav2vec} and HuBERT \cite{hsu2021hubert}. "Text" denotes using the text SSL model, such as mBART \cite{chipman2022mbart} and LLaMA2 \cite{touvron2023llama}. $^\dag$ uses the multimodal model jointly trained with speech and text as initialization. LST-7B and LST-13B indicate that we use the 7B and 13B versions of LLaMA2 as backends to perform a complete two-stage training.}
\label{table_main}
\end{table*}

\subsection{Experimental Setups}
Our models consist of three modules. For the \emph{speech frontend}, we use the CTC finetuned Wav2vec 2.0 large model\footnote{https://dl.fbaipublicfiles.com/fairseq/wav2vec/wav2vec\_vox\_\\960h\_pl.pt}, which is pre-trained with 53.2k hours of untranscribed speech from LibriVox \cite{kearns2014librivox}, fine-tuned on the 960h of transcribed speech from Librispeech, and on pseudo-labels \cite{xu2021self}. We discard the CTC projection head, and utilize the last encoder output as speech representation. The dimension of the speech representation is 1024. The \emph{adapter module} consists of two parts: the length adapter and the modality adapter. The length adapter includes two 1-dimensional convolutional layers with a kernel size of 5, a stride size of 2, padding of 2, and a hidden dimension of 1024.. The modality adapter is just a simple linear layer to connect the length adapter and LLM backend. We choose LLaMA2\footnote{https://huggingface.co/meta-llama} \cite{touvron2023llama} as the \emph{LLM backend}, as its effectiveness has been demonstrated in several open-source leaderboard. Its embedding dim is 4096.

We use the AdamW \cite{loshchilov2017decoupled} optimizer. We use cosine learning rate decay and the warmup ratio is set to 0.03. The learning rate is 2e-3 and 2e-5 in the first and second stage, respectively. We train the first stage for 6 epochs. In the second stage, the parameter of the LLM model is trainable. Because of the large capacity and its strong memory power, we find the model can quickly overfit after one epoch, bringing a small performance boost. Hence, we only train one epoch for the second stage. The batch size in both stage is set to 128. Due to time constraints, we do not perform an extensive hyperparameter search.

In the first and second stage, we save checkpoints every 1000 and 100 steps, respectively. During inference, we use the final checkpoint for evaluation, have found that averaging multiple checkpoints does not yield better results. We use beam search with a beam size of 4. We use sacreBLEU \cite{post2018call} to compute case-sensitive detokenized BLEU scores for a fair comparison with previous works. All models are trained on 16 Nividia Tesla V100 GPUs with the help of ZeRO training strategy \cite{rajbhandari2020zero}. The training for first and second stages takes 23h and 26h respectively.

\begin{table*}[]
\centering
\begin{tabular}{l|ccccc}
\hline
\multicolumn{1}{c|}{Strategy} & \multicolumn{1}{c}{Stage1(ASR)} & \multicolumn{1}{c}{Stage1(ST)} & \multicolumn{1}{c}{Stage2} & \multicolumn{1}{c}{BLEU(stage1)} & \multicolumn{1}{c}{BLEU(stage2)} \\ \hline
(a)    & $\times$     & $\checkmark$     &  $\times$       & 24.91       & $-$                        \\
(b)    & $\times$     & $\times$         &  $\checkmark$   & $-$         & 19.93                    \\
(c)    & $\checkmark$ & $\times$         &  $\checkmark$   & $-$         & 28.73                  \\
(d)    & $\times$     & $\checkmark$     &  $\checkmark$   & 24.91       & 29.15                  \\ \hline
\end{tabular}
\caption{LST-7B model performance on MuST-C en-de tst-COMMON set under different training strategy. Stage1(ASR) and Stage1(ST) indicate that we use the ASR and ST datasets in the first stage, respectively. We report the BLEU scores for first and second stage.}
\label{table_stra}
\end{table*}

\subsection{Baseline Models}
We mainly compare the proposed method with some strong end-to-end ST baselines including: Fairseq-ST \cite{wang2020fairseq}, STAST \cite{liu2020bridging}, SATE \cite{xu-etal-2021-stacked}, TDA \cite{du2022regularizing}, Chimera \cite{6acdf73253e64b7eb908c3a90c13e2d9}, ConST \cite{ye2022cross}, FCCL \cite{Zhang2023ImprovingST}, STEMM \cite{fang-etal-2022-stemm}, M$^3$ST \cite{cheng2023m}, CRESS \cite{fang2023understanding}, MSP \cite{zhang2023improving}, STPT \cite{tang2022unified}, SpeechUT \cite{zhang2022speechut}. These models have significantly contributed to advances in the E2E-ST task in the past, achieving advanced performance. We compare these methods to demonstrate the effectiveness of our proposed method. 

\section{Results}
We compare the performance of our method with existing end-to-end methods in Table \ref{table_main}. LST-7B has already surpassed most previous models. Remarkably, LST-13B achieves new state-of-the-art performance. TDA \cite{du2022regularizing} is trained from with the help of triangular
decomposition agreement strategy, in which the transcription in source language is used. LST-13B shows a +3.2 higher BLEU score in En-De language pair, by solely connecting the single modality pretrained models, without employing any additional mature training methods or transcription, indicating that training from scratch is not the optimal choice. MSP \cite{zhang2023improving} also uses the single modality pretrained models, initializing the encoder and decoder of the E2E-ST model with Wav2vec 2.0 and the fine-tuned mBART decoder on machine translation task, respectively. Then, it is further end-to-end fine-tuned using LNA \cite{li2021multilingual}. In contrast, LST is simple and flexible, and has the ability to combine advanced research results in single modality. By simply scaling the LLM backend from 7B to 13B, the model performance can be significantly improved (34.05-$>$35.22). Overall, LST achieves superior performance with its streamlined structure and training strategy. However, it is important to note that our method is orthogonal to theirs as they focus on advanced training methods and are model-agnostic. We plan to explore avenues for integrating these approaches in future investigations.

\section{Analysis}
\subsection{Effects of Training Strategy}
As previously mentioned, we employed a two-stage training approach. In this section, we examine the impact of different training stages on the final performance. The model performance after the first and second stage training in various strategies is reported in Table \ref{table_stra}. We observe the following phenomena:

\noindent\textbf{(1) The two stages of training are indeed necessary; the first and second stages alone will yield poor results.} First and second stage alone will produce poor results. Results from Table \ref{table_stra} demonstrate that the final performance of strategies (a) and (b) is inferior to that of strategy (d). When considering the first stage alone, it can be viewed as performing prompt tuning on LLM, with the upper limit depending on LLM's own reasoning capability. There are many capabilities in LLM that are not related to specific downstream tasks and take up a portion of the model capacity. Therefore, the effect is limited without adjusting LLM parameters. However, we observe that without the first stage training, the performance of the second stage training alone is even worse than that of the first stage training alone. We speculate that this is because the input representation to LLM differs significantly from the text representation space without modality adjustment. In such cases, updating LLM parameters may disrupt its own understanding ability, resulting in poor performance.

\noindent\textbf{(2) In the first stage of training, using direct ST data yields better results compared to ASR data, even if the ASR data consists of a total of 960h while the ST data only comprises 400h.} As shown Table \ref{table_stra}, strategy (d) slightly outperforms strategy (c). Previous works often employ ASR data for initial feature alignment. Their focus is on making the model more generic. However, in scenarios where the downstream task is known, aligning features directly using the downstream task data proves to be sufficient.

\begin{table}[]
\centering
\begin{tabular}{l|ll}
\hline
Models               & BLEU(stage1) & BLEU(stage2) \\ \hline
Small                & 16.60        & 23.51        \\
Large                & 17.12        & 23.97        \\
CTC fine-tuned Large & 24.91        & 29.15        \\ \hline
\end{tabular}
\caption{LST-7B model performance on MuST-C en-de tst-COMMON set under different speech frontend. All three models are from publicly released checkpoints, and we have not made any further adjustments.}
\label{table_ssl}
\end{table}

\subsection{Effects of Speech frontend}
The speech frontend is used to extract semantic representation from input speech signal. In this section, we analyze the effects of the speech frontend. We compared three versions of Wav2vec 2.0: the small, the large and the CTC fine-tuned large model. The CTC fine-tuned large model is obtained by finetuning the Wav2vec 2.0 large model using Librispeech data on ASR task. As shown in Table \ref{table_ssl}, the large version exhibits greater capability in extracting representations compared to the small version, resulting in slightly better outcomes. However, incorporating the CTC fine-tuned Wav2vec 2.0 large model as the frontend significantly enhances the model's performance, indicating linguistic supervision contained in the speech frontend plays a key role in the final performance. We hypothesize that this might be due to the relatively lightweight nature of the adapter we use, limiting its capacity for modal adjustment, while the disparity between the transformed speech representation and the representation space of text embedding is substantial. For the model fine-tuned on the ASR task, the output representation of the last layer already closely approximates the text representation, rendering a lightweight adapter sufficient. In future work, we will explore more intricate structures.

\begin{table}[]
\centering
\begin{tabular}{l|l|ll}
\hline
Size                & Models     & BLEU(stage1) & BLEU(stage2) \\ \hline
\multirow{3}{*}{7B} & LLaMA      & 23.66        & 27.88        \\
                    & Vicuna 1.3 & 24.18        & 27.36        \\
                    & LLaMA2     & 24.91        & 29.15        \\ \hline
13B                 & LLaMA2     & 26.42        & 30.39        \\ \hline
\end{tabular}
\caption{LST-7B model performance on MuST-C en-de tst-COMMON set under different LLM backend. Vicuna 1.3\footnote{https://huggingface.co/lmsys/vicuna-7b-v1.3} \cite{vicuna2023} is a finetuned model based on LLaMA \cite{thoppilan2022lamda} using the sharegpt data, showing better instruction-following ability.}
\label{table_llm}
\end{table}

\subsection{Effects of LLM backend}
The LLM serves as the backend for generating text translations in the target language based on speech features. However, the selection criteria for the LLM backend are still an open question. In this section, we conduct relevant analysis and present the results in Table \ref{table_llm}. Based on the obtained results, we draw the following conclusions:

\noindent\textbf{(1) Supervised instruction fine-tuning (SFT) is not essential for the LLM backend.} When Vicuna is employed as the backend, the first stage demonstrates significantly better performance compared to LLaMA, while the second stage exhibits slightly worse performance. This discrepancy can be attributed to the fact that only the adapter is trainable in the first stage, thus relying on the reasoning ability of the LLM, which is enhanced after SFT. However, in the second stage, both the adapter and the LLM parameters are trainable. The LLM after SFT may exhibit a certain bias, leading to adverse effects.

\noindent\textbf{(2) The overall performance of the model on the E2E-ST task hinges on the foundation LLM.} LLaMA2, benefiting from a larger training dataset, possesses a stronger foundational capability, resulting in significantly better performance than LLaMA.

\noindent\textbf{(3) The larger the LLM model, the higher the final performance achieved on the E2E-ST task.} LLaMA2 13B outperforms the 7B version substantially due to its enhanced foundational capability.

\begin{table*}[]
\centering
\begin{tabular}{ll}
\hline
\multicolumn{1}{l|}{Models}                 &      \\ \hline
\multicolumn{2}{c}{CASE1(ted\_1404\_23)}                        \\ \hline
\multicolumn{1}{l|}{\multirow{2}{*}{Ref}}   & src: Lights, sounds, solar panels, motors — everything should be accessible. \\
\multicolumn{1}{l|}{}                       & tgt: Lichter, Töne, Solarelemente, Motoren — alles sollte verfügbar sein. \\ \hline
\multicolumn{1}{l|}{ConST}                  & tgt: \textcolor{red}{\underline{Licht}}, Geräusche, Solarpanele, Motoren, alles sollte zugänglich sein. \\ \hline
\multicolumn{1}{l|}{LST}                    & tgt: Lichter, Geräusche, Solarzellen, Motoren, alles sollte zugänglich sein. \\ \hline
\multicolumn{2}{c}{CASE2(ted\_1404\_1)}                          \\ \hline
\multicolumn{1}{l|}{\multirow{4}{*}{Ref}}   & src: Eight years ago when I was at the Media Lab, I started exploring this idea of how to put the power of \\
\multicolumn{1}{l|}{}                       & \quad \ \ \ engineers in the hands of artists and designers.     \\
\multicolumn{1}{l|}{}                       & tgt: Vor acht Jahren war ich am Media Lab und ich begann diese Idee zu erforschen, wie man die Macht  \\
\multicolumn{1}{l|}{}                       & \quad \ \ \ der Ingenieure in die Hand von Künstlern und Designern legen könnte.     \\ \hline
\multicolumn{1}{l|}{\multirow{2}{*}{ConST}} & tgt: Vor acht Jahren, als ich im Media Lab war, begann ich, diese Idee zu erforschen, wie man die Macht \\
\multicolumn{1}{l|}{}                       & \quad \ \ \  \textcolor{red}{\underline{von}} Ingenieuren in die Hände von Künstlern und Designern \textcolor{red}{\underline{legt}}.   \\ \hline
\multicolumn{1}{l|}{\multirow{2}{*}{LST}}   & tgt: Vor acht Jahren, als ich am Media Lab war, begann ich, diese Idee zu erforschen, wie man die Macht  \\
\multicolumn{1}{l|}{}                       & \quad \ \ \ der Ingenieure in die Hände von Künstlern und Designern legen kann.     \\ \hline
\end{tabular}
\caption{En-De test cases that generated from the ConST and our LST-13B model. The \textcolor{red}{\underline{red underlined text}} indicates incorrect or inaccurate translations.}
\label{table_case}
\end{table*}

\begin{figure}[]
\centering
\includegraphics[width=0.4\textwidth]{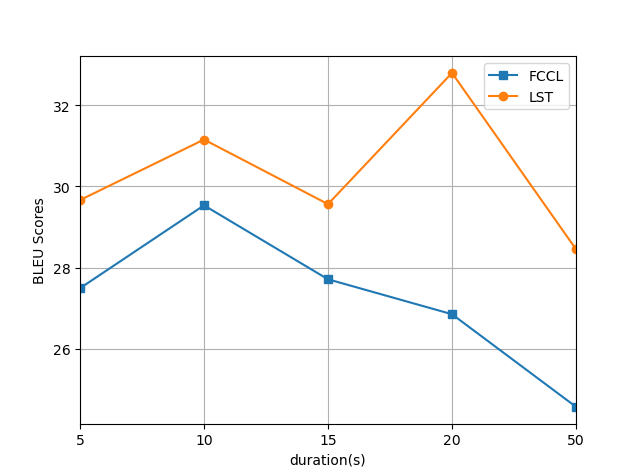} 
\caption{BLEU scores on different groups of source speech in terms of their duration.}
\label{figure_length}
\end{figure}

\subsection{Results in Different Speech Lengths}
We further evaluate the performance of LST on long speech translation, where conventional vanilla E2E-ST models often exhibit poor results. We believe that the LLM's robust reasoning capabilities can play a crucial role in improving long speech translation. To investigate this, we conduct experiments on 5 separate groups according to the duration of the source speech in tst-COMMON sets. Figure XX show the BLEU scores. FCCL \cite{Zhang2023ImprovingST} represents the previous E2E-ST model. Notably, we observe that the performance curve of LST consistently surpasses that of ConST by a significant margin. On long speech, the improvement is even greater. Particularly, in the last two groups of speech, our LST obtains 5.94 and 3.88 BLEU points improvement over FCCL, respectively. Overall, these consistent improvements across all groups in terms of the varying durations of the source speech, highlight the effectiveness of LST in translating long speech signals. We will conduct further analysis in the following case study.

\begin{figure}[htbp]
  \centering

  \begin{subfigure}[b]{0.23\textwidth}
    \includegraphics[width=\textwidth]{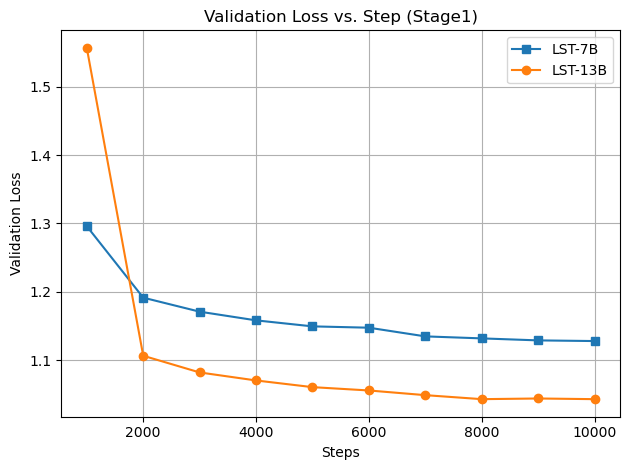}
    \caption{stage1}
    \label{fig:subfig1}
  \end{subfigure}
  \hfill
  \begin{subfigure}[b]{0.23\textwidth}
    \includegraphics[width=\textwidth]{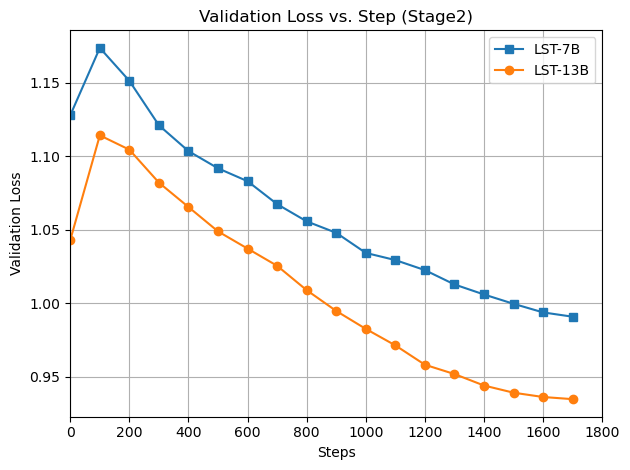}
    \caption{stage2}
    \label{fig:subfig2}
  \end{subfigure}

  \caption{Model learning dynamics on MuST-C en-de dataset.}
  \label{fig:subfigures}
\end{figure}

\subsection{Case study}
In this section, we present several cases generated by LST to compare its performance with the previous end-to-end model, ConST \cite{ye2022cross}. To ensure a fair comparison with prior works, we utilize the same test cases reported in their paper. The results are summarized in Table \ref{table_case}.
In the first case, ConST incorrectly translates "Lights" as "Licht," while LST accurately translates it as "Lichter". In the second case, which is a longer sample, both models initially provide grammatically correct translations. However, ConST exhibits more translation errors in the later part. For instance, both "der" and “legt” are  incorrect translations. In contrast, LST correctly translates these words, resulting in a more accurate overall translation than ConST. These observations further highlight the long speech translation capabilities of LST

\subsection{Learning dynamic}
Figure \ref{fig:subfigures} depict the learning dynamics of the model in the first and second stages. 

\noindent\textbf{In the first stage,} as the number of training steps increases, we observe a consistent decrease in the model's loss on the validation set. Furthermore, the 13B model outperforms the 7B model significantly. However, it is important to note that at the beginning of training, the validation loss of the 13B model is higher than that of the 7B model. We hypothesize that this discrepancy arises due to the larger number of parameters in the 13B model. Consequently, employing the same learning rate for both models might not be suitable. A potential solution to address this issue could be to simply reduce the learning rate. This adjustment may potentially resolve the disparity in the validation loss between the two models For time reasons, we do not make further learning rate adjustments and longer training.

\noindent\textbf{In the second stage,} the learning dynamics exhibit a similar pattern as observed in the first training stage. However, there is an increase in the validation loss during the early phase of training. This can primarily be attributed to the transition from a learning rate of 0, which may not be suitable. An excessively small learning rate can potentially undermine the knowledge acquired during the first stage. Moving forward, we intend to explore more effective methods for adjusting the learning rate.
 
\section{Relate work}
\noindent\textbf{E2E-ST.} End-to-end ST \cite{weiss2017sequence,berard2018end,sperber2019attention} has gained increasing attention due to its structural advantages. However, the modal complexity and data scarcity make it difficult to train such a model well. To overcome the modal complexity, various training techniques have been applied to assist training, including pre-training \cite{bansal2019pre,stoian2020analyzing,wang2020curriculum}, multi-task learning \cite{weiss2017sequence,liu2020synchronous}, meta-learning \cite{indurthi2019data}, contrastive learning \cite{ye2022cross,ouyang2022waco,Zhang2023ImprovingST}, knowledge distillation \cite{liu2019end,gaido2020end,gaido2020knowledge,inaguma-etal-2021-source,zhao-etal-2021-mutual}, training sampling \cite{fang2023understanding}. Moreover, the encoder component of E2E-ST models often carries a heavy burden, as it needs to extract both acoustic features and semantic information. To alleviate this burden, some models decouple the encoder into an acoustic encoder and a semantic encoder \cite{dong2021listen}. Additionally, the CTC-based shrink mechanism \cite{liu2020bridging,xu-etal-2021-stacked} to reduce the length of speech features, thereby effectively improving model performance. To overcome the data scarcity, methods like data augmentation \cite{bahar-etal-2019-using,lam-etal-2022-sample,zhao-etal-2023-generating,fang2023back,zhang2023dub}, non-parametric knowledge distillation \cite{zhang2023decoupled}, network structure design \cite{zhang2022revisiting}, mixup \cite{fang-etal-2022-stemm,cheng2023m}, have been explored to improve the utilization of existing E2E-ST data. 

Recently, researchers \cite{li2021multilingual,gallego2021upc,zhang2023improving} have achieved significant improvements in baseline performance by combining self-supervised speech models with finetuned mBART \cite{chipman2022mbart}. In this approach, the speech SSL model serves as the encoder, while the decoder of mBART, after fine-tuning on the MT task, is used as the decoder. The entire model is further finetuned end-to-end using the ST data. However, this method does not fully exploit the potential of advanced LLMs in NLP, and its scalability is limited. 

In this paper, we explore how to improve the performance of E2E-ST by leveraging the capabilities of state-of-the-art LLMs. Our approach allows for more flexibility in selecting the speech SSL frontend and LLM backend, enabling prompt integration of research advancements in the single modality domain. This flexible combination demonstrates robust scalability.

\noindent\textbf{Multimodal LLM}. LLM has made rapid progress in natural language processing (NLP), demonstrating remarkable understanding capabilities \cite{thoppilan2022lamda,touvron2023llama}. It has shown that language can play a wider
roles: universal interface for a general-purpose assistant \cite{liu2023visual}. Thus, a natural step is multimodal LLM, which other modalities are uniformly transformed into text embedding space, and LLM is used as an interface to perform various downstream tasks according to instructions.
Zhang et al. \cite{zhang2023speechgpt} discretize speech representations and further train LLaMA \cite{thoppilan2022lamda} to develop SpeechGPT, which demonstrates an impressive capacity to follow multi-modal human instructions. LTU \cite{gong2023listen}, LLaSM \cite{Shu2023llasm} and X-LLM \cite{chen2023x} are also open-source efforts that enable LLaMA to utilize speech inputs, laying the foundation for building open-source multimodal LLMs.

However, although multimodal LLM in the field of speech shows the characteristics of being able to listen and speak, its performance in specific downstream tasks is usually insufficient \cite{chen2023x}. Our preliminary exploration shows that it performs poorly on complex tasks, such as E2E-ST, a cross-language and cross-modal translation task, falling far behind single-modal models. In contrast, this paper primarily focuses on specializing LLM to excel at the E2E-ST task.

\section{Conclusion}
In this paper, we propose LST, a large multimodal model that excels at E2E-ST task. Experiments on MuST-C datasets demonstrate the effectiveness of our proposed method, which outperforms the previous models and achieves new state-of-the-art. 

This project is still ongoing and there are several directions for further exploration: (1) Using more complex adapter. The current adapter is lightweight, resulting in limited modality adjustment capabilities. A more sophisticated adapter may result in better performance. (2) Using better LLM. Due to time and resource constraints, only 13B LLM model is used as the back end. It can be expected that using a stronger language model, LST will gain more powerful capabilities. (3) In combination with other methods. Because of the simple structure and training strategy, LST can actually be seen as a strong baseline and we will explore the combination with other approaches.

\bibliography{aaai24}

\end{document}